\documentclass[a4paper,12pt]{article}

\usepackage[noblocks]{authblk}
\usepackage[left=1.5cm,right=1.5cm,top=1.5cm,bottom=2cm]{geometry}

\title{\textbf{Probabilistic Model of Visual Segmentation}}

\author[1]{Jonathan Vacher}
\author[2]{Pascal Mamassian}
\author[1]{Ruben Coen-Cagli} 

\affil[1]{\small Albert Einstein College of Medicine,
Dept. of Systems and Comp. Biology,
10461 Bronx, NY, USA}
\affil[2]{
Laboratoire des Syst\`emes Perceptifs, D\'epartement d'\'Etudes Cognitives,
\'Ecole Normale Sup\'erieure, PSL Research University, CNRS,
29 rue d'Ulm, 75005 Paris, France
}

\date{}

\usepackage{hyperref}

\usepackage{mystyle}
\usepackage{wrapfig}

\usepackage{nicefrac}       
\usepackage{hhline}

\begin{document}

\maketitle

\begin{abstract}
Visual segmentation is a key perceptual function that partitions visual space and allows for detection, recognition and discrimination of objects in complex environments. The processes underlying human segmentation of natural images are still poorly understood. In part, this is because we lack segmentation models consistent with experimental and theoretical knowledge in visual neuroscience. 
Biological sensory systems have been shown to approximate probabilistic inference to interpret their inputs. This requires a generative model that captures both the statistics of the sensory inputs and expectations about the causes of those inputs. Following this hypothesis, we propose a probabilistic generative model of visual segmentation that combines knowledge about 1) the sensitivity of neurons in the visual cortex to statistical regularities in natural images; and 2) the preference of humans to form contiguous partitions of visual space.
We develop an efficient algorithm for training and inference based on expectation-maximization and validate it on synthetic data. Importantly, with the appropriate choice of the prior, we derive an intuitive closed--form update rule for assigning pixels to segments: at each iteration, the pixel assignment probabilities to segments is the sum of the evidence (\ie local pixel statistics) and prior (\ie the assignments of neighboring pixels) weighted by their relative uncertainty.
The model performs competitively on natural images from the Berkeley Segmentation Dataset (BSD), and we illustrate how the likelihood and prior components improve segmentation relative to traditional mixture models. Furthermore, our model explains some variability across human subjects as reflecting local uncertainty about the number of segments. Our model thus provides a viable approach to probe human visual segmentation.

\end{abstract}

\section{Introduction}
\label{sec:intro}

\paragraph{Segmentation in computer vision and computational neuroscience} Image segmentation is a long standing topic in computer vision (for review see \eg~\cite{pal1993review,morel2012variational,dey2010review}) with applications ranging from environment-machine interaction (\eg autonomous cars, exploring robots, UAVs) to computer-aided diagnosis (\eg medical imaging, video surveillance). Because of the difficulty to gather large sets of human-segmented images, segmentation remains largely an unsupervised problem. Thus, deep neural networks have had limited success, with the exception of the more constrained problem of semantic segmentation~\cite{badrinarayanan2017segnet,ronneberger2015u,chen2018deeplab,long2015fully} which involves assigning each pixel of the image to a class out of set of pre-specified classes (e.g. 'car', 'pedestrian', 'road', etc.). While this approach is successful at producing perceptually meaningful image segmentations of the training classes, it does not generalize to arbitrary images. Traditional approaches to unsupervised segmentation used graph--based methods~\cite{shi2000normalized,felzenszwalb2004efficient} which view an image as a graph to be partitioned. Approaches based on feature similarity are also common and rely on the intuition that the human visual system tends to group together features that share the same properties~\cite{alpert2012image,ning2010interactive,puzicha1997non}. From a complementary perspective, segmentation is often reduced to the problem of contour detection~\cite{MartinFTM01,arbelaez2011contour}, and related algorithms approached state of the art performance~\cite{maninis2016convolutional,zhao2015segmenting}. Yet, these algorithms do not achieve human level performance and, when they fail, they often produce segmentations that are not perceptually meaningful. 

Here we propose instead a different approach, based on the hypothesis that biological sensory systems perform approximate probabilistic Bayesian inference to extract meaning from noisy and ambiguous sensory signals~\cite{pouget2013probabilistic,knill1996perception,knill2004bayesian,dayan2001theoretical,kersten2004object}. Segmentation is a prime example of inference on ambiguous inputs: the pixels of an image do not contain sufficient information for labeling them as grouped or segmented always with certainty. Therefore, here we address segmentation explicitly as a problem of probabilistic inference. Relevant work in computer vision includes algorithms based on probabilistic mixture models extended to include Bayesian priors that favor grouping by proximity~\cite{nikou2010bayesian}, e.g. via the distance dependent Chinese Restaurant Process \cite{ghosh2011spatial}, the spatially dependent Pitman-Yor process~\cite{sudderth2009shared}, and the Location Dependent Dirichlet Process (LDDP)~\cite{sun2017location}, to restrain the search to partitions composed of contiguous components. Although promising, a main limitation of those approaches is their use of ad--hoc image descriptors and over--simplified assumptions about their statistics. This is in contrast with the hypothesis above, which implies that efficient segmentation in humans and other primates should rely on an accurate model of the statistics of the features to which neurons are sensitive in natural images, as we discuss next. 
 
\paragraph{Segmentation in primate vision} The segmentation of an image into individual objects and the integration of elementary features to build these objects are two competing and challenging tasks that visual systems have to solve simultaneously. Much work in human visual psychophysics has focused on grouping and segmentation in simple artificial displays, such as boundaries between oriented textures, aligned edges forming a long contour embedded in randomly oriented distractors, simple shapes grouped by common motion patterns, and illusory edges and figures revealing perceptual expectations of simplicity and good continuation (for review see ~\cite{wagemans2012century}). This work is mirrored by electrophysiology in non-human primates, which has revealed that neurons in early and mid-level areas of the visual cortex are sensitive to those segmentation cues~\cite{roelfsema2006cortical,slllito1995visual,cavanaugh2002nature,li2006contour,pasupathy2015neural}. As such, visual segmentation can be viewed as a feedforward process that is strongly refined and modulated by feedback and lateral connections~\cite{neri2017object,li1999contextual} also allowing for fast peripheral detection of specific objects~\cite{boucart2016finding}. However, much less is known about segmentation of complex, natural images. Two notable exceptions are studies of contour grouping \cite{geisler2009contour,sigman2001common} and figure-ground judgments~\cite{fowlkes2007local}, that showed how these processes rely largely on low-level image cues, thus suggesting a direct link between natural image statistics and segmentation.

\paragraph{Natural image statistics and cortical representations} Natural images are characterized by the power-law decay of their power spectrum, and an abundance of oriented edges due, \eg, to occlusions and texture boundaries~\cite{torralba2003statistics,aaponatural}. Indeed, several unsupervised algorithms trained on natural images recover localized, oriented filters that are comparable with receptive fields (RFs) of neurons in primary visual cortex (V1)~\cite{olshausen1996emergence,bell1997independent}, suggesting that V1 neurons represent over-complete wavelet coefficients of natural images. Importantly, images have also higher-order statistical structure, most prominently the nonlinear dependence between pairs of wavelet coefficients: the variance of one coefficient depends on the magnitude of the other coefficient~\cite{wainwright2000scale}. These dependencies are captured well by Gaussian Scale Mixture (GSM) models, which assume that a latent global random variable (representing, \eg contrast) scales a set of local Gaussian variables (representing the intensity of edges) thus introducing statistical coordination between wavelet coefficients~\cite{wainwright2000scale}. GSM models have been used in practical applications, such as denoising~\cite{portilla2003image,rakvongthai2010complex}, contrast enhancement~\cite{lyu2008nonlinear} and texture synthesis~\cite{theis2012mixtures}. Furthermore, V1 neurons are sensitive to the dependencies described by the GSM~\cite{schwartz2001natural}, and probabilistic inference in mixtures of GSMs (mGSM; details in Section 2) explains nonlinear properties of V1 (\eg contextual modulation~\cite{coen2009statistical,Coen-Cagli2012} and temporal adaptation~\cite{snow2016specificity}) and achieves state-of-the-art predictions of V1 responses to natural images~\cite{coen2015flexible}. Intuitively, the dependencies represented within the GSM reflect the properties that are shared within an object and that differ from other objects, thereby offering a natural strategy to segment the image. Therefore, here we build on, and substantially extend, this class of models.

\paragraph{Contributions} We will combine the strength of the probabilistic approach to segmentation with accurate statistical models of the image features to which visual cortex is sensitive. We will focus on V1 because it has been studied extensively and it is strongly modulated by segmentation cues~\cite{slllito1995visual}, but our approach could be extended hierarchically to include features that reflect the selectivity of neurons in higher cortical areas, such as texture descriptors~\cite{freeman2013functional} and hidden layers of deep networks~\cite{yamins2014performance}. The statistics of V1--like image features (localized oriented filters) are accurately described by mGSMs, and V1 neurons are sensitive to the statistical similarity between features inside the RF and those in the surround, in natural images~\cite{coen2015flexible}. Building on those results, here we propose a segmentation algorithm based on statistical grouping. Specifically, we associate each pixel to a feature vector consisting of pixel values (colors) and wavelet coefficients (orientations, scales) which are jointly modeled by a mGSM, regularized by a prior that encourages grouping of nearby areas of the visual field. First, we derive an efficient algorithm for learning and inference, validate it on synthetic data, and show analytically that it satisfies a key requirement of correct probabilistic inference: namely, when computing the probability that a pixel is assigned to a segment, it combines the evidence (i.e. local pixel statistics) with the prior (i.e. the assignments of neighboring pixels) weighted by their relative uncertainty. Next, we apply our algorithm to natural images, and show that it performs competitively with state of the art methods, and additionally that it captures uncertainty in human segmentations. Interestingly, we also demonstrate that the computation of uncertainty often provides an accurate contour map, thus offering a link between contour-based segmentation and region-based segmentation. 

\paragraph{Notations} We use the following notation. Integers $D, N$ and $K$ denote respectively the dimension, number of samples, and number of classes/labels. A random variable is denoted by a capital letter $X$. The probability density function of $X$ is denoted $\pr{X}$ while $x_n$ denotes a sample. The set $\LL$ is the image pixel lattice and in our framework any sample $x_n$ is associated to a pixel location $l_n \in \LL$. The set $\Delta^K$ represents the $K$-dimensional simplex. Finally, ``pdf'' stands for ``probability density function'' and unless stated differently, all random variables are considered on the vector space $\RR^D$.

\section{Segmentation inference by expectation-maximization}
\label{sec:algo-em}

First, we recall in Definition~\ref{def:gsm-student} the pdf of the multivariate Student-t distribution.
\begin{defn}
\label{def:gsm-student}
A random vector $X$ is a multivariate Student-t random vector if it has the following pdf
\eql{\label{eq:mult-stud}
\pr{X}(x;\alpha) = \pr{X}(x;\nu,\mu,\Sigma) = \frac{\Gamma(\nicefrac{(D+\nu)}{2})\al^{\nicefrac{D}{2}}}{\Gamma(\nicefrac{\nu}{2})(2\pi)^{\nicefrac{D}{2}}\vert \Sigma \vert^{\nicefrac{1}{2}} (1+\nicefrac{1}{\nu}d_{\Sigma}(x,\mu)^2)^{\nicefrac{(D+\nu)}{2}}}
}
where $\Gamma(v) = \int_{\RR_+} u^{v-1} e^{-u} \d u$ is the Gamma function and $d_{\Sigma}(x,\mu)  = \sqrt{(x-\mu)^T \Sigma^{-1}(x-\mu)}$ is the Mahalanobis distance. For ease of notation, we denote $\alpha=(\nu,\mu,\Sigma)$.
\end{defn}
The multivariate Student-t is specific case of GSM, obtained from the ratio of a Gaussian vector and a Chi real random variable. In the limit of infinite $\nu$, the Student-t converges to a Gaussian distribution with mean $\mu$ and covariance $\Sigma$.  

Here we consider a mixture of Student-t distributions with mixing probabilities that depend on the pixel location $l \in \LL$
\eq{
\pr{X}(x; \al) =  \sum_{k=1}^{K} p_{k}(l)  \pr{X^{(k)}}(x; \al_k)
}
where $\alpha=((\alpha_k)_k)$ are the Student-t parameters and for $l \in \LL, p(l)=(p_1(l),\dots,p_K(l)) \in \Delta^K$ are the mixing probabilities. To train our model, we adopt the Expectation-Maximization (EM) approach which consists in completing the sample data with their class. Therefore, we consider $((x_n,c_n))_{n\in \{1,\dots, N\}}$ where for all $n \in \{1,\dots, N\}$, $c_n \in \{1,\dots,K\}$ is the class of $x_n$. For ease of notation, we denote the mixing probabilities by $p_{n,k}=p_k(l_n)$ in the following. As such, with a non-informative (uniform) prior, the learned mixing probabilities $p_{n,k}$ converge towards the posterior probabilities $\cpr{C_n}{X_n}(k|x_n)$. In order to avoid such an over-fitting, we consider a regularizing prior on the mixing probabilities parametrized by location and scale $(m,s)$ that we denote $\pr{P}(\cdot;m,s)$. Learning is achieved using the maximum \textit{a posteriori} (MAP) estimator
\eql{\hat \theta = \uargmin{\boldsymbol{\theta}}  \ell\left((x_n,c_n)_{n}; \boldsymbol{\theta} \right)  
}
with $\boldsymbol{\theta}=(\boldsymbol{p,\alpha})=((p_{n,k})_{n,k},(\al_k)_k)$ and where $\ell$ is the negative log-posterior function 
\eql{
\ell\left((x_n, c_n)_{n}; \boldsymbol{\theta} \right) =  - \mathop{\sum_{n=1}^{N,K}}_{k=1} \one_k(c_n) \ln\left( p_{n,k}  \pr{X^{(k)}}(x_n; \al_k)  \right) +  \ln\left( \pr{P}(p_{n,k};m,s)\right).
}
with $\one_k(c)=1$ if $c=k$ and $\one_k(c)=0$ otherwise.

The expectation step is standard for mixture distributions and only requires the computation of the posterior assignment of pixels to segments, at the $t$-th iteration 
\eql{
\tau_{n,k}^{(t)} \eqdef \EE(\one_k(C_n)\vert x_n,\boldsymbol{\theta}^{(t)}) = \cpr{C_n}{X_n,\boldsymbol{\Theta}}(k|x_n,\boldsymbol{\theta}^{(t)}) = \frac{p_{n,k}^{(t)}  \pr{X^{(k)}}(x_n; \al_k^{(t)})}{\sum_{i=1}^{K} p_{n,i}^{(t)}  \pr{X^{(i)}}(x_n; \al_i^{(t)})}.
} 

Then, the maximization step involves minimizing the expectation of the completed-data negative log-posterior 
\eql{
\label{eq:Q-update}
Q(\boldsymbol{\theta}; \boldsymbol{\theta}^{(t)}, (x_n)_n) = -  \mathop{\sum_{n=1}^{N,K}}_{k=1}    \tau_{n,k}^{(t)}  \Big[   \ln(p_{n,k}) + \ln\left( \pr{X^{(k)}}(x_n; \al_k) \right) \Big]  + \ln\left( \pr{P}(p_{n,k};m_n,s_n)\right).
}
The first and last terms of Equation~\eqref{eq:Q-update} only depend on $p_{n,k}$ while the middle term only depends on $\al_k$, therefore the update of the model parameters $\al_k$ is independent from the update of the prior probabilities $p_{n,k}$. The update of the model parameters for Student-t mixture follows from the work of Peel and McLachlan~\cite{peel2000robust} which is summarized in Proposition~\ref{prop:eq-extemum}.
\begin{prop}
\label{prop:eq-extemum}
At an extremum point of $Q$, the following equations hold for all $k\in \{1,\dots,K\}$ 

\vspace{2mm}
\noindent
\begin{minipage}{0.4\linewidth}
\eql{\label{eq:mean}
\mu_k  = \frac{\sum_{n=1}^N \tau_{n,k}^{(t)}\omega_{n}(\alpha_k) x_n }{\sum_{n=1}^N \tau_{n,k}^{(t)}\omega_{n}(\alpha_k)},
}
\end{minipage}
\hspace{1mm}
\begin{minipage}{0.58\linewidth}
\eql{\label{eq:cov}
\Sigma_k  = \frac{\sum_{n=1}^N \tau_{n,k}^{(t)}\omega_{n}(\alpha_k) (x_n-\mu_k)(x_n-\mu_k)^T }{\sum_{n=1}^N \tau_{n,k}^{(t)}},
}
\end{minipage}
\vspace{4mm}
\eql{\label{eq:power}
\ln(\nicefrac{\nu_k}{2})-\psi(\nicefrac{\nu_k}{2}) = \kappa_k(\alpha_k),
} 
where $\omega_{n}(\alpha_k) =  \frac{\nu_k+D}{\nu_k+d_{\Sigma_k}(x_n,\mu_k)^2}$, $\psi(x) = \frac{\Gamma^\prime(x)}{\Gamma(x)}$ is the digamma function and
\eql{
\kappa_k(\alpha_k) = \ln(\frac{\nu_k+D}{2}) - \psi(\frac{\nu_k+D}{2}) - 1 -\frac{1}{N_k^{(t)}} \sum_{n=1}^N  \tau_{n,k}^{(t)} (\ln(\omega_{n}(\alpha_k) ) - \omega_{n} (\alpha_k) ) 
}
with $N_k^{(t)}=\sum_{n=1}^N \tau_{n,k}^{(t)} $.
\end{prop}
The four equations of Proposition~\ref{prop:eq-extemum} have no closed form solutions. However, in such form, Equations~\eqref{eq:mean},~\eqref{eq:cov} and~\eqref{eq:power} can be solved recursively by using their estimates from the last iteration of EM  $(\boldsymbol{\nu^{(t)},\mu^{(t)},\Sigma^{(t)}})$ in the right hand terms. Specifically, Equation~\eqref{eq:power} requires an additional numerical step to update $\boldsymbol{\nu}$ (Newton-Raphson)~\cite{peel2000robust}.
 
As indicated above, we use a prior $\pr{P}(\cdot;m,s)$ parametrized by location and scale $(m,s)$. Previously proposed priors relied on a specific elaborations of the first term of Equation~\eqref{eq:Q-update} which is required to guarantee that the prior probabilities sum to $1$~\cite{sun2017location,nikou2010bayesian} and often lead to solve a non-linear system of equations to update the prior probabilities. In contrast, Dirichlet and Logit-Normal distributions are defined on the simplex, therefore appropriate as prior distributions. The Logit-Normal distribution~\cite{atchison1980logistic} is directly parametrized by location and scale, however the update equations are also non-linear. The Dirichlet distribution was previously used to enforce spatial dependence directly~\cite{ghosh2011spatial} or in combination with a Gaussian process~\cite{sun2017location}. We propose instead a specific parametrization that has the considerable advantage of leading to a linear regularizing equation, that also satisfies a key requirement of well-calibrated probabilistic inference, i.e. weighting by reliability, as explained below. First, we recall in Definition~\ref{def:dirichlet} the pdf of the Dirichlet distribution.

\begin{defn}
\label{def:dirichlet}
A random vector $P\in\Delta^K$ is a Dirichlet random vector if it has the following pdf
\eql{\label{eq:dirichlet}
\pr{P}(p;a) = \frac{\Gamma\left(\sum_{k=1}^K a_k \right)}{\prod_{k=1}^K \Gamma(a_k)} \prod_{k=1}^K p_k^{a_k-1}
}
where $a=(a_1,\dots,a_K) \in \RR_+^K$.
\end{defn}
We use the following location and scale parametrization 
\eq{
a = \frac{m}{s^2} + \mathbf{1}
}
where $m = (m_1,\dots,m_K) \in \Delta^K$ and $\mathbf{1}$ is a K-dimensional vector of $1$. This parametrization leads to update equations given in Proposition~\ref{prop:prior}.
\begin{prop}
\label{prop:prior}
At an extremum point of $Q$, the following equations holds for all $k\in \{1,\dots,K\}$ 
\eql{
\label{eq:prior-update}
p_{n,k} = \frac{s_n^2\tau_{n,k}^{(t)}+m_{n,k}}{s_n^2+1}.
}
\end{prop}
\begin{proof}
Add the Lagrange multiplier $\lambda$ associated to the constraint $\sum_k p_{n,k}=1$ to $Q$ in Equation~\ref{eq:Q-update}. Compute the derivative and set $\lambda$ such that the constraint is met. 
\end{proof}
The prior update corresponds to a weighted mean between the posterior $\tau_{n,k}^{(t)}$ and the location $m_{n,k}$. The scale $s_n$ is the uncertainty on the location vector $m_{n} \in \Delta^K$. If it is high, the update relies more on the posterior, if it is low the update relies more on the location vector $m_n$. Regularization is achieved by choosing $(m_n,s_n)$ to be the weighted mean and variance of nearby pixels posterior $(\tau_{m,k}^{(t)})_{m \in \Nn_n}$ where $\Nn_n$ is a neighborhood of pixel $n$. Using the notations $m_{n,k} = m_k(l_n), \tau_{n,k}^{(t)} = \tau_k^{(t)}(l_n) \text{ and } p_{n,k} = p_k(l_n)$, we have

\begin{minipage}{0.3\linewidth}
\eql{\label{eq:loc}
m_k(l_n) = G \ast \tau_{k}^{(t)}(l_n) 
}
\end{minipage}
\hspace{1mm}
\begin{minipage}{0.65\linewidth}
\eql{\label{eq:scale}
s(l_n)^2 = \frac{1}{K(1-G\ast G(0))} \sum_{k=1}^K G \ast {\tau_{k}^{(t)}}^2(l_n) - m_k(l_n)^2
}
\end{minipage}
where $\ast$ is the discrete 2D convolution. In practice, $G$ is a Gaussian kernel with width $\sigma$ which controls the area of spatial averaging.

\section{Results}
\label{sec:results}

In this section, first, we validate the model on artificial data and show that it reliably estimates underlying spatial class probabilities. Then, we evaluate the performance in natural image segmentation on the BSD 500~\cite{arbelaez2011contour} in comparison with simple Student-t mixture and Gaussian mixture with and without Dirichlet prior. We show few segmentation examples that illustrate the typical behavior of each algorithm. Last, we discuss segmentation uncertainty in humans using the BSD to exemplify how mixture models will be relevant to further study on human segmentation of natural images.

\paragraph{Validation on artificial data}
We synthesized two color images with 3 segments and different uncertainty level in the prior assignment of pixels to segments (see Figure~\ref{fig:artif-im}, first row). Color vectors of pixels are Student-t distributed with different means, covariances and degrees of freedom per segment. We run our inference algorithm with kernel width $\sigma$ equal to $5.25$ and $10.25$ respectively for low and high uncertainty (Figure~\ref{fig:artif-im} left and right). Inferred prior probability maps closely match the true probability maps (average absolute error of $0.023$ and $0.027$ respectively). As expected, reconstruction is often very good when the Student-t distributions for each segment are well separated but is impaired when they are overlapping (\eg equal means, not shown).

\myfigureH{\includegraphics[scale=0.35]{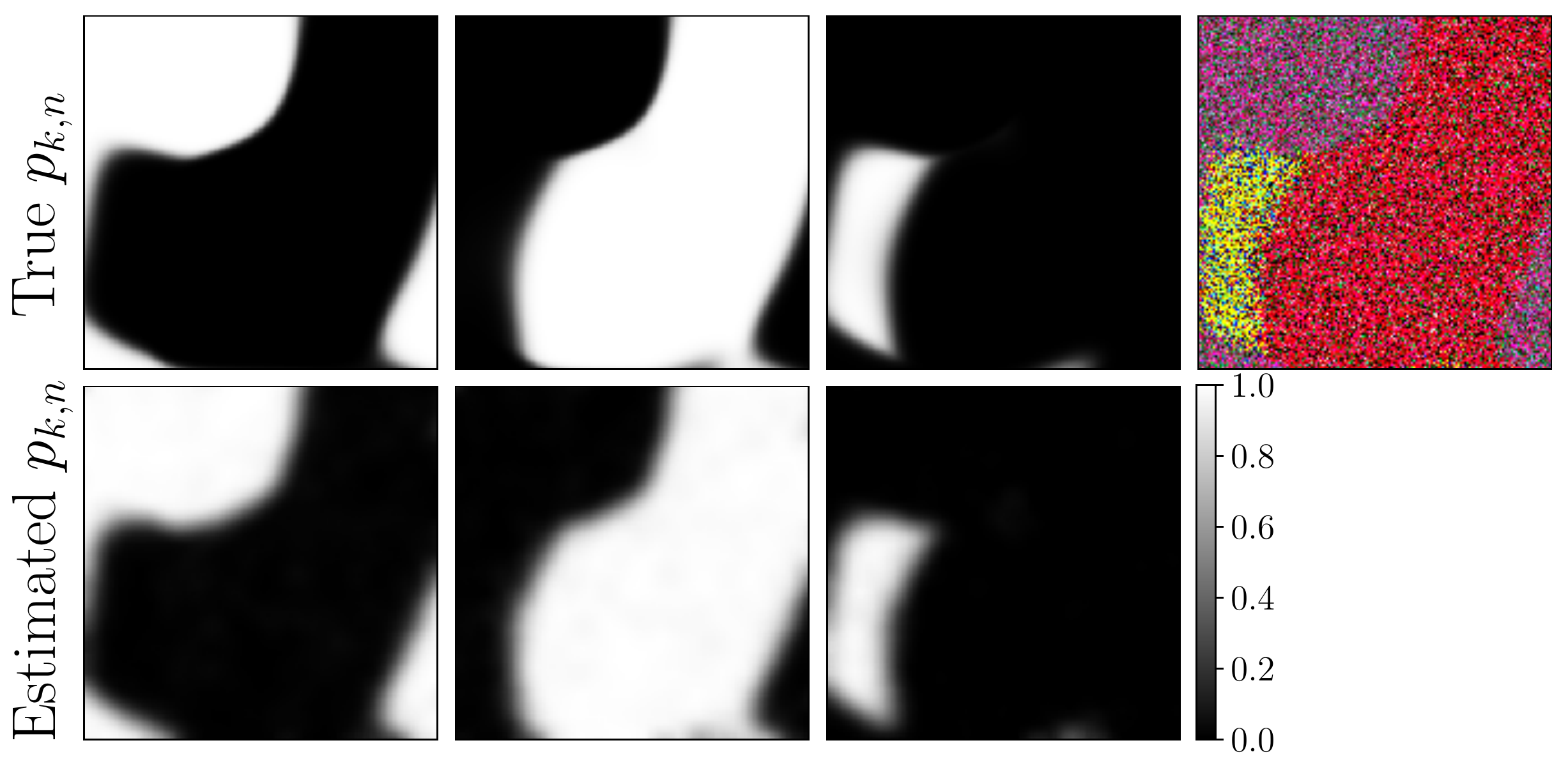}
\includegraphics[scale=0.35]{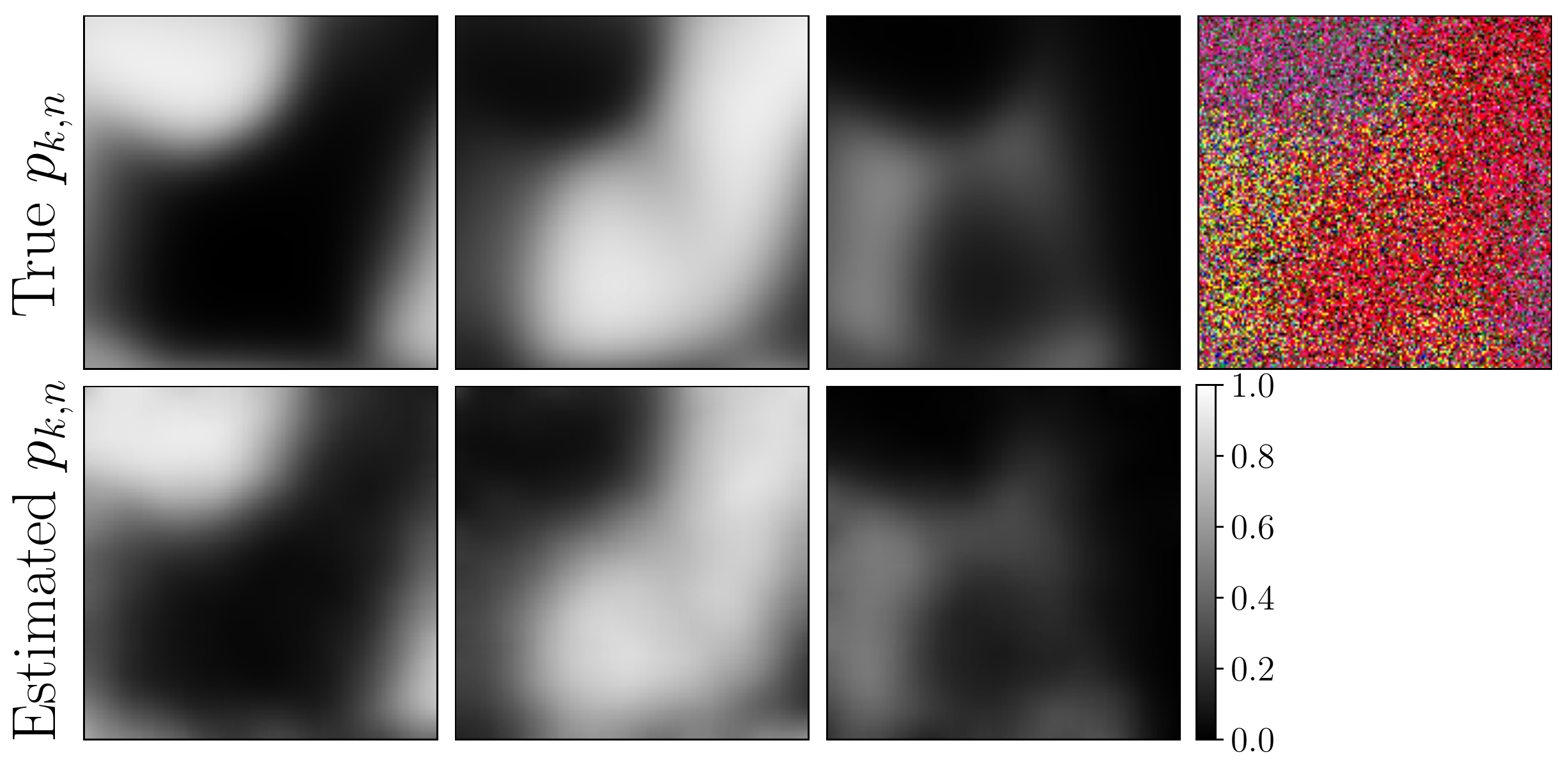}}{Validation on artificial images. Data are generated from prior probabilities $p_{n_k}$ and Student-t components in the 3-dimensional color space. Left: low uncertainty \ie prior probability are close to $0$ or $1$ at each pixels. Right: high uncertainty \ie prior probabilities are balanced between $0$ and $1$ across the image.}{fig:artif-im}

\newcolumntype{C}{>{\centering\arraybackslash\hsize=.5\hsize}X}
\newcolumntype{K}{>{\centering\arraybackslash}X}
\newcolumntype{L}{>{\raggedright\arraybackslash}X}
\begin{wraptable}{R}{0.55\textwidth}
    \centering
    \begin{tabularx}{0.55\textwidth}{|L | K | C | C | }
        \hline
        & aRI (RI)   & $F_b$     & $F_{op}$    \\ \hline
		Stud.-t + Dir. & \bf{0.461} (\bf{0.796})        & \bf{0.674}        & \bf{0.209}        \\ \hline
        Stud.-t & 0.414 (0.780)        & 0.639        & 0.172         \\ \hline
        Gauss. + Dir. & 0.246 (0.669)       & 0.475        & 0.168         \\ \hline
        Gauss. & 0.171 (0.587)         & 0.382        & 0.145         \\ \hhline{|=|=|=|=|}  
        Human & 0.700 (0.878)        & 0.803        & 0.556         \\ \hhline{|=|=|=|=|}        
        COB (\cite{maninis2016convolutional}) & n/a (n/a)       & 0.793      & 0.419   	\\ \hline
        Ncut (\cite{cour2005spectral}) &    n/a (0.80)      & 0.641        & 0.213         \\ \hline
    \end{tabularx}
    \caption{Results on the BSD 500 for the three scores referred in text. Bold scores are the best among the 4 tested probabilistic algorithms.}
   	\label{tab:results}
\end{wraptable}

\paragraph{Segmentation performance}
We tested our inference algorithm on the BSD database of natural images with segmentations performed by human subjects \cite{MartinFTM01,arbelaez2011contour}. We compare the Student-t mixture and the Gaussian mixture models with and without Dirichlet prior, using three standard scores: the adjusted Rand Index (aRI), the F-score for boundaries ($F_b$) and the F-score for objects and parts ($F_{op}$)~\cite{pont2013measures} (see Table~\ref{tab:results}). We also show the Rand Index (RI) because it is commonly used in previous works (see~\cite{pont2013measures} and references therein). Yet, RI is likely to suffer from bias and did not score well with meta-measures~\cite{pont2013measures} while aRI has not been evaluated with meta-measures. We run each algorithm for $K \in \{2,\dots,9\}$ and for each score we choose the value of $K$ that maximizes the score. We associate to each pixel of the image a vector of color features and V1-like features consisting of 4 orientations at 1 scale (or spatial frequency) on which we perform PCA to remove noise (\ie we keep $99.9\%$ of the variance). For the Dirichlet prior, we use a Gaussian kernel with width $\sigma=4.25$. Among the 4 tested probabilistic algorithms, the Student-t mixture with Dirichlet prior performs best in all scores. In both Student-t and Gaussian models, the prior improves all three scores. Strikingly, the simple Student-t mixture performs better than the Gaussian mixture with Dirichlet prior demonstrating the importance of using a model that is in line with natural image statistics. As a comparison, in Table~\ref{tab:results} we also show human scores, the scores for a state of the art algorithm called COB~\cite{maninis2016convolutional}, and the classical Ncut algorithm~\cite{maninis2016convolutional}. Our algorithm's performance is competitive but still falls short of the state of the art. Our main focus here was not on maximizing performance, but rather providing a probabilistic framework to study human segmentation, however there are a few clear directions to improve performance (see also next subsection). First, different from COB and Ncut, our algorithm is not specifically crafted for contour detection; specifying that information in our model could improve the resulting segmentation. Second, our implementation used oriented filters at a single spatial scale, but it could be readily extended to multiple scales. Third, as explained above we focused only on V1--like oriented filters, whereas COB relies also on higher--order features. Including those features in our framework is an important direction for future work. Lastly, note that our algorithm compares least favorably on the $F_{op}$ score: this is expected, because good performance on the $F_{op}$ requires some high--level knowledge of objects, which is absent from our framework. 

\paragraph{Segmentation examples} To assess qualitatively the segmentations of the 4 tested algorithms, we show in Figure~\ref{fig:nat-im} one example with a good score and another example with a bad score. For both models the Dirichlet prior reduces residual noise (isolated pixels classified differently from surrounding pixels). The sensitivity of the Gaussian mixture to outliers is visible in both examples: one segment dominates all others. This is because, during learning, one estimated covariance will tend to have large eigenvalues to account for outliers and will therefore tend to cover a large part of the dataset.
\myfigureH{
\includegraphics[scale=0.48]{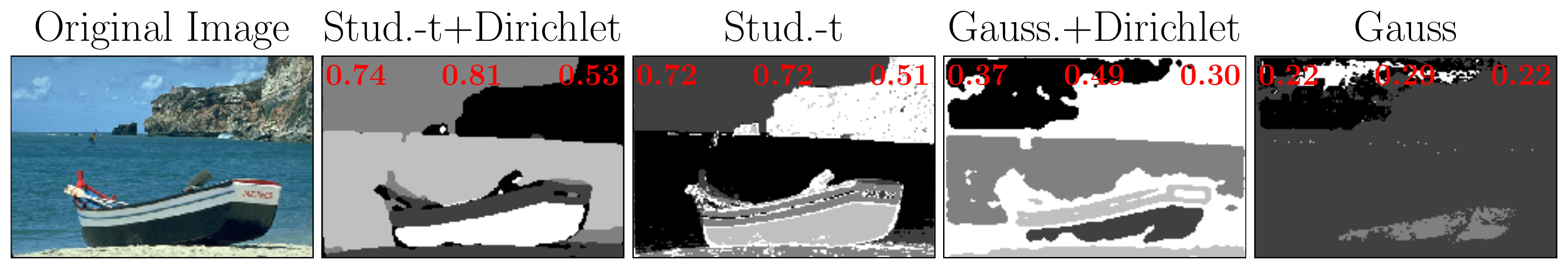}\\
\includegraphics[scale=0.48]{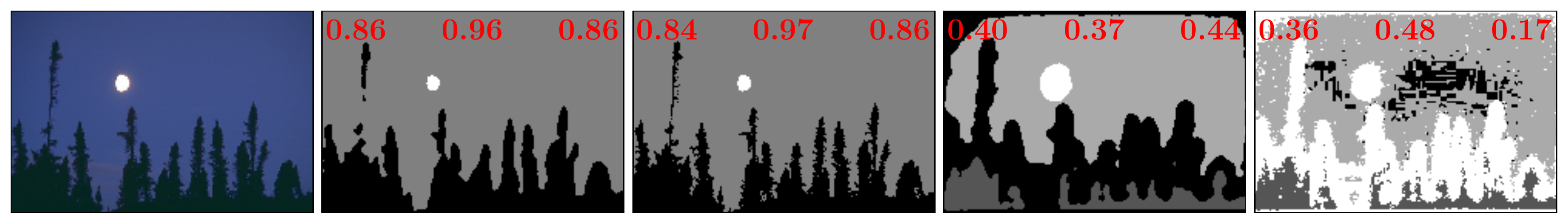}\\
\includegraphics[scale=0.48]{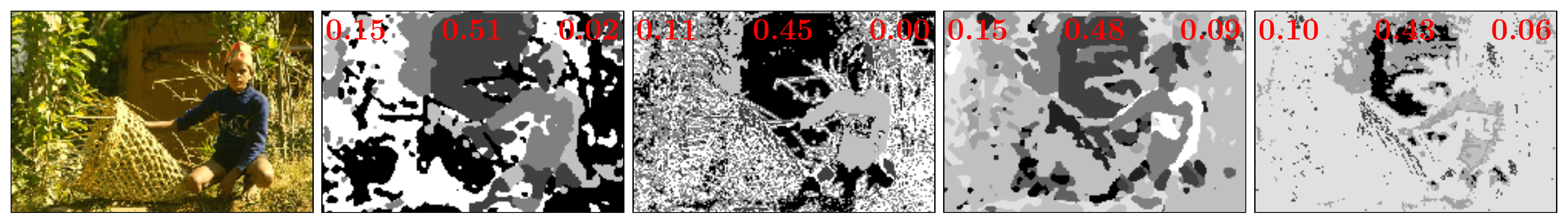}\\
\includegraphics[scale=0.48]{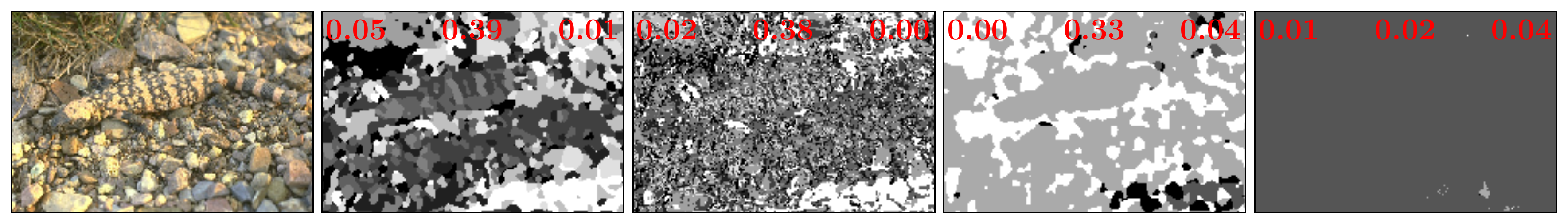}}{Segmentation examples for the four tested algorithms. First row: good scores.  Bottom row: bad scores. The three scores (aRI, $F_c$, $F_{op}$) are indicated in red.}{fig:nat-im}
This can be understood because natural images have heavy-tailed distributions, \ie they contain many outliers for a Gaussian distribution. Such a sensitivity to outliers is partly corrected when using the Dirichlet prior and completely avoided when using the Student-t mixture. The Student-t mixture, besides being robust to outliers, is also a specific case of GSM which is known as a good statistical model of wavelet coefficients of natural images. Interestingly, the Student-t mixture captures segments that are relevant to human vision (grouping principle~\cite{wagemans2012century}): in addition to mixture components capturing large areas that form objects, it also often groups the pixels belonging to boundaries surrounding objects into a separate mixture component. The boundary grouping is partly reduced when using the Dirichlet prior. Interestingly, an alternative approach could be to retain some explicit 'boundary' mixture components by using more structured kernels $G$ to parametrize the boundary prior (e.g. an elongated rather than circular kernel). Last, we observe that the failure example contains higher frequency content related to rough textures. Therefore as suggested above, including lower frequency and/or hierarchical features could help improving the segmentation maps in those cases.

\paragraph{Segmentation uncertainty} We next tested whether our framework can account for the variability of human segmentations, as is required for guiding future quantitative studies of human perceptual segmentation of natural images. We first analyzed variability across subjects in the BSD 500. We found that variability in the number of segments per image approximately follows a Weber-Fechner law, \ie the standard deviation of the number of segments scales with the number of segments (Figure~\ref{fig:cor-entropy} left). Note that BSD 500 was collected under weakly controlled conditions, therefore this variability across subjects may reflect multiple causes and confounding factors, including that different subjects interpreted the instructions differently. However, the observed Weber--Fechner law indicates that variability across subjects is structured, and it may partly reflect uncertainty of the individual subjects, driven by image properties. For instance, the segmentation of pixels near boundaries between similar textures is expected to be more uncertain than for pixels inside a uniform texture. 
\myfigure{
\includegraphics[scale=0.4]{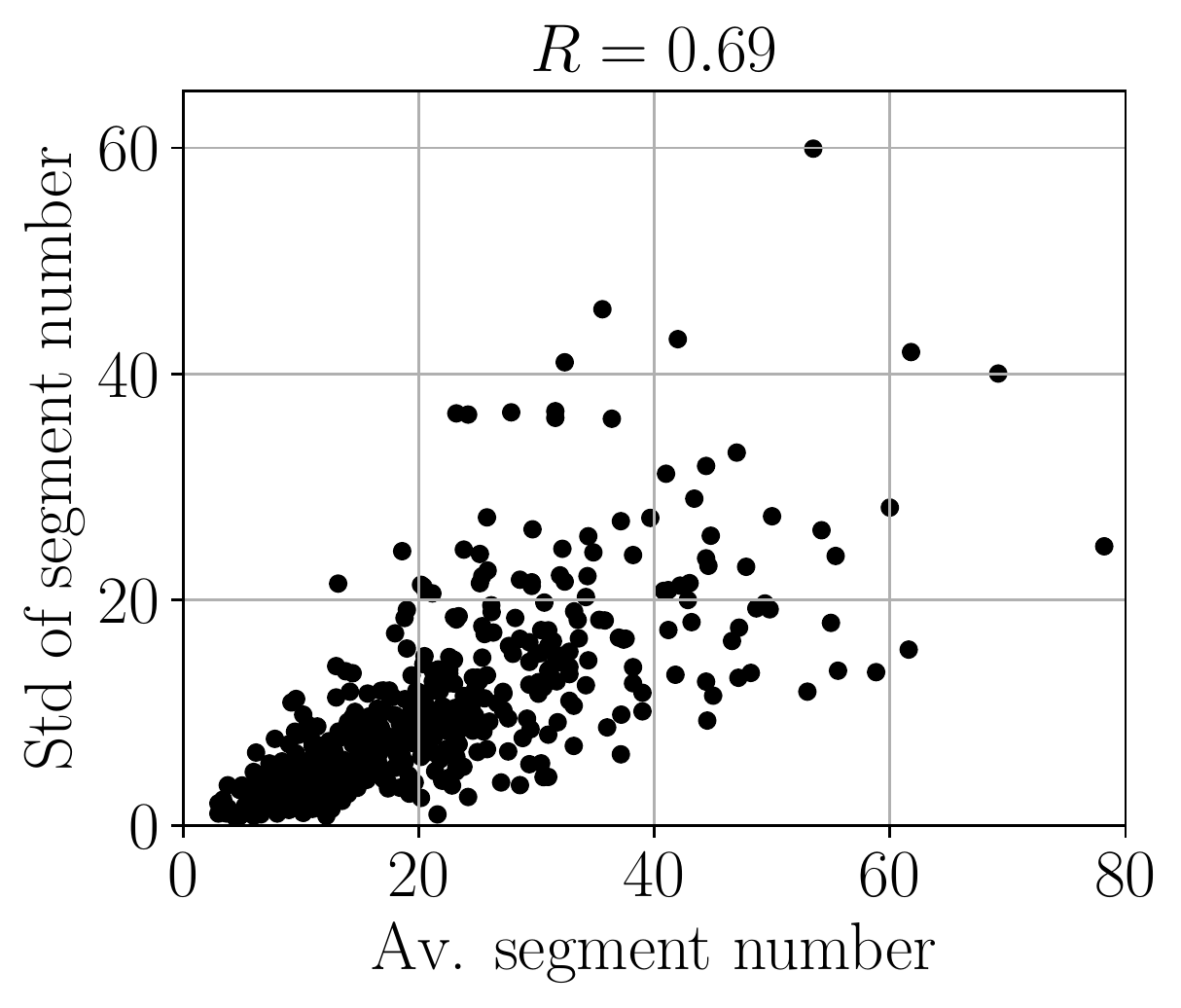}
\includegraphics[scale=0.4]{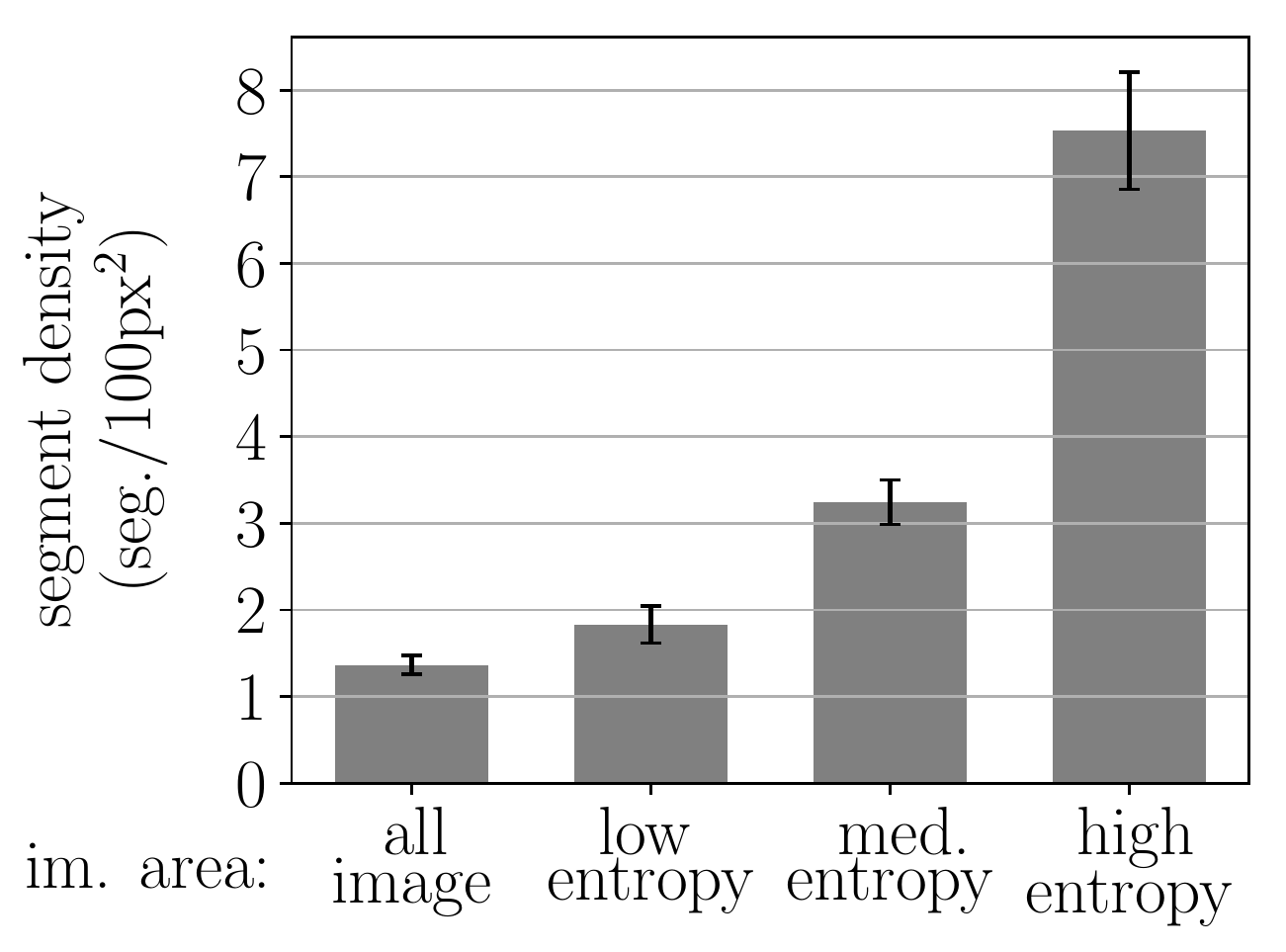}
\includegraphics[scale=0.4]{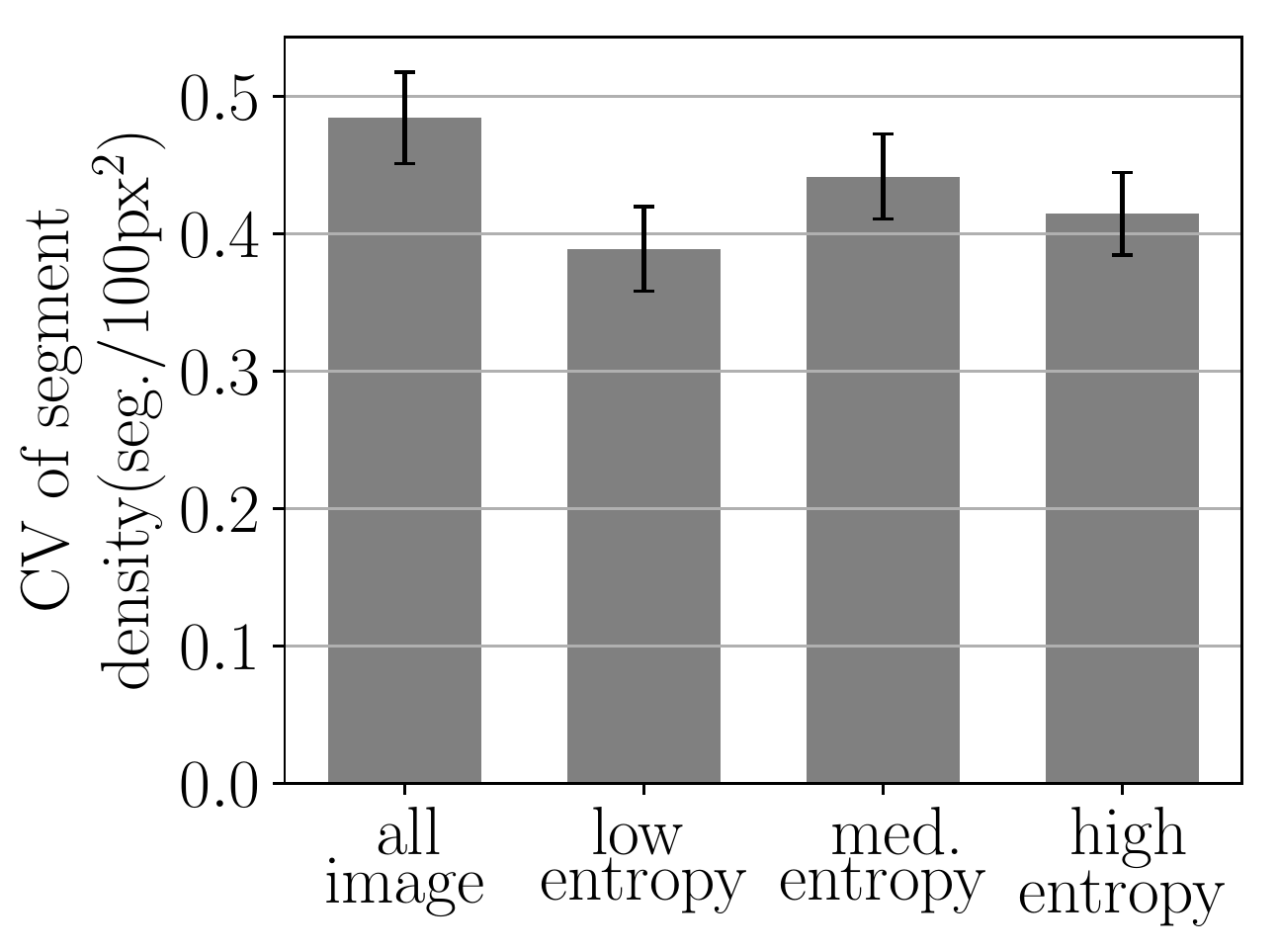}
}{Left: Weber-Fechner law in segment number. Pearson correlation coefficient is indicated on top. Middle and right: Model entropy correlates with segment number uncertainty.}{fig:cor-entropy}
\begin{wrapfigure}{r}{0.5\textwidth}
\begin{center}
\includegraphics[scale=0.36]{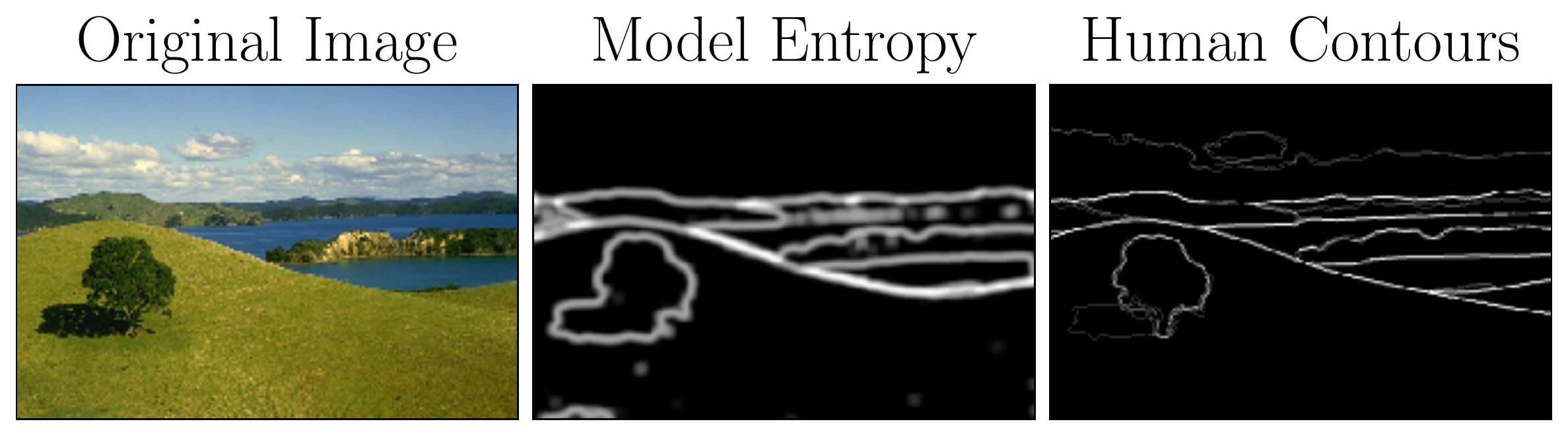}\\
\includegraphics[scale=0.36]{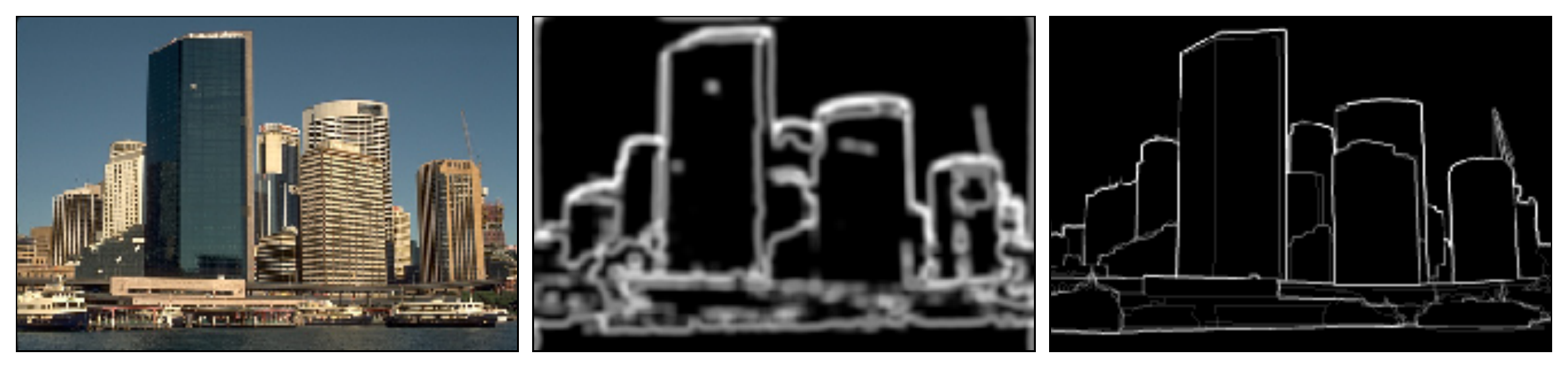}
\end{center}
\vspace{-5mm}
\caption{Entropy is high at contour location.}
\label{fig:entropy}
\end{wrapfigure}
This posited link between uncertainty and variability is a specific instantiation of our hypothesis that the brain performs approximate probabilistic inference. This is because the hypothesis requires that probability distributions, and not simply point estimates, are represented by neural activity and reflected in perceptual judgments; uncertainty is a key aspect of the represented probability (i.e. its width) and larger uncertainty will result in larger variability of repeated perceptual judgments given a fixed visual input~\cite{pouget2013probabilistic,fiser2010statistically}. Importantly, our algorithms provide a probabilistic output and therefore allowed us to compute the predicted segmentation uncertainty at each pixel location $l \in \LL$. Specifically, we computed the entropy of segment attribution (i.e. of $p_{k}(l)$). Interestingly, our result highlights a good match between human contours and the boundaries of the algorithm's segmentation maps, because those boundaries correspond to probability transition between segments (between $0$ and $1$) and therefore to high categorical entropy. Thus, consistent with our intuition, entropy correlates well with human drawn contours (Figure~\ref{fig:entropy}).

The finding that entropy is highest at boundaries also leads to a specific prediction about human variability: areas with higher average entropy under the proposed mixture model should correspond to areas with a larger mean and standard deviation of the number of segments in human maps. We found that the human data agreed well with the prediction (Figure~\ref{fig:cor-entropy} middle, and examples in Figure~\ref{fig:entropy-ex}). The effect on the standard deviation was fully explained by that on the mean, i.e. we did not observe a correlation between the predicted entropy level and the coefficient of variation of the number of human segments (Figure~\ref{fig:cor-entropy} right).
 
\myfigureH{\includegraphics[scale=0.254]{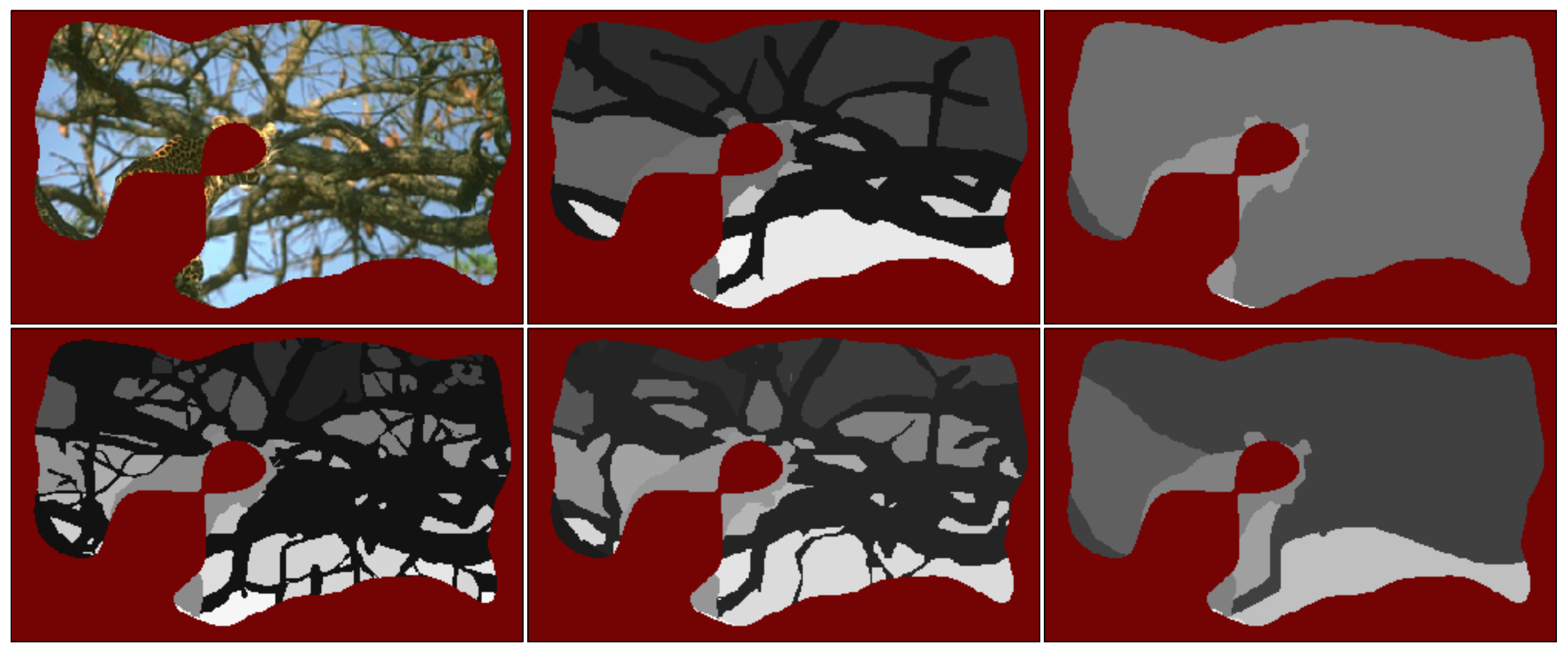}
\includegraphics[scale=0.23]{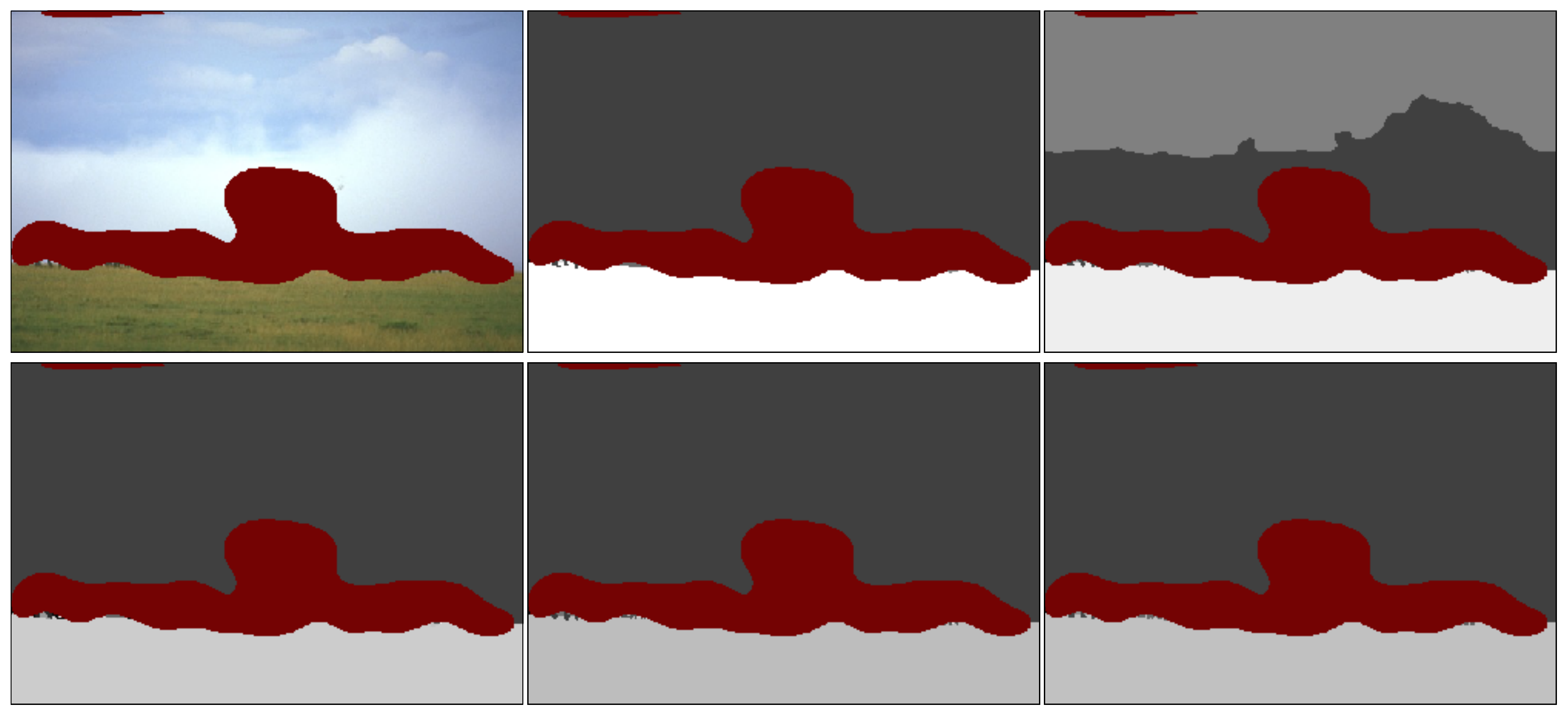}
\includegraphics[scale=0.254]{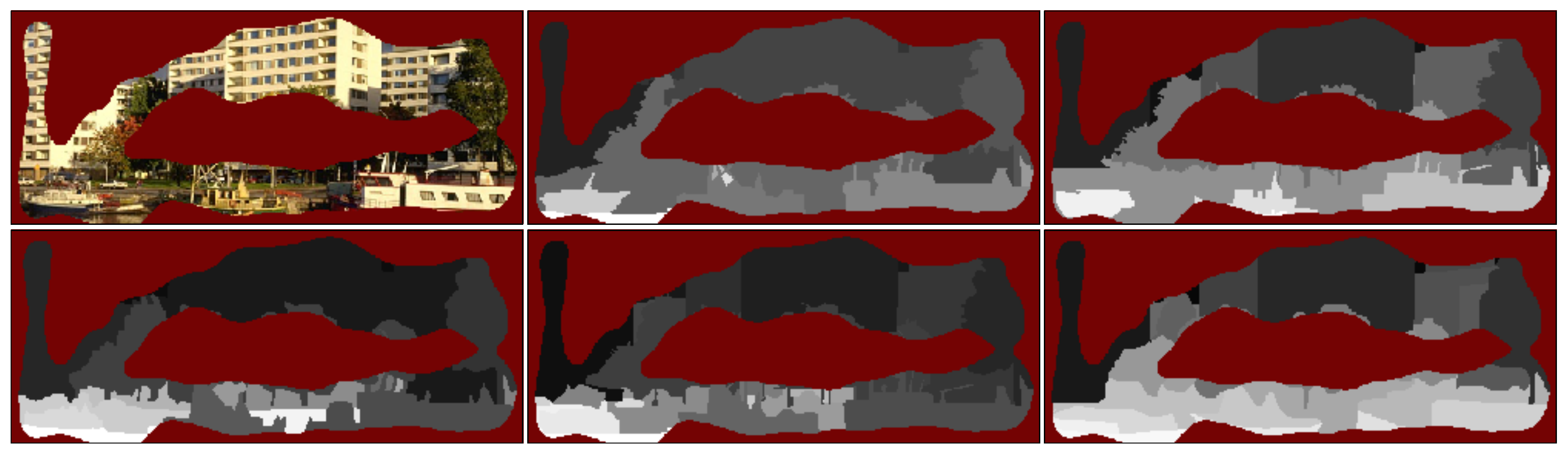}
\includegraphics[scale=0.23]{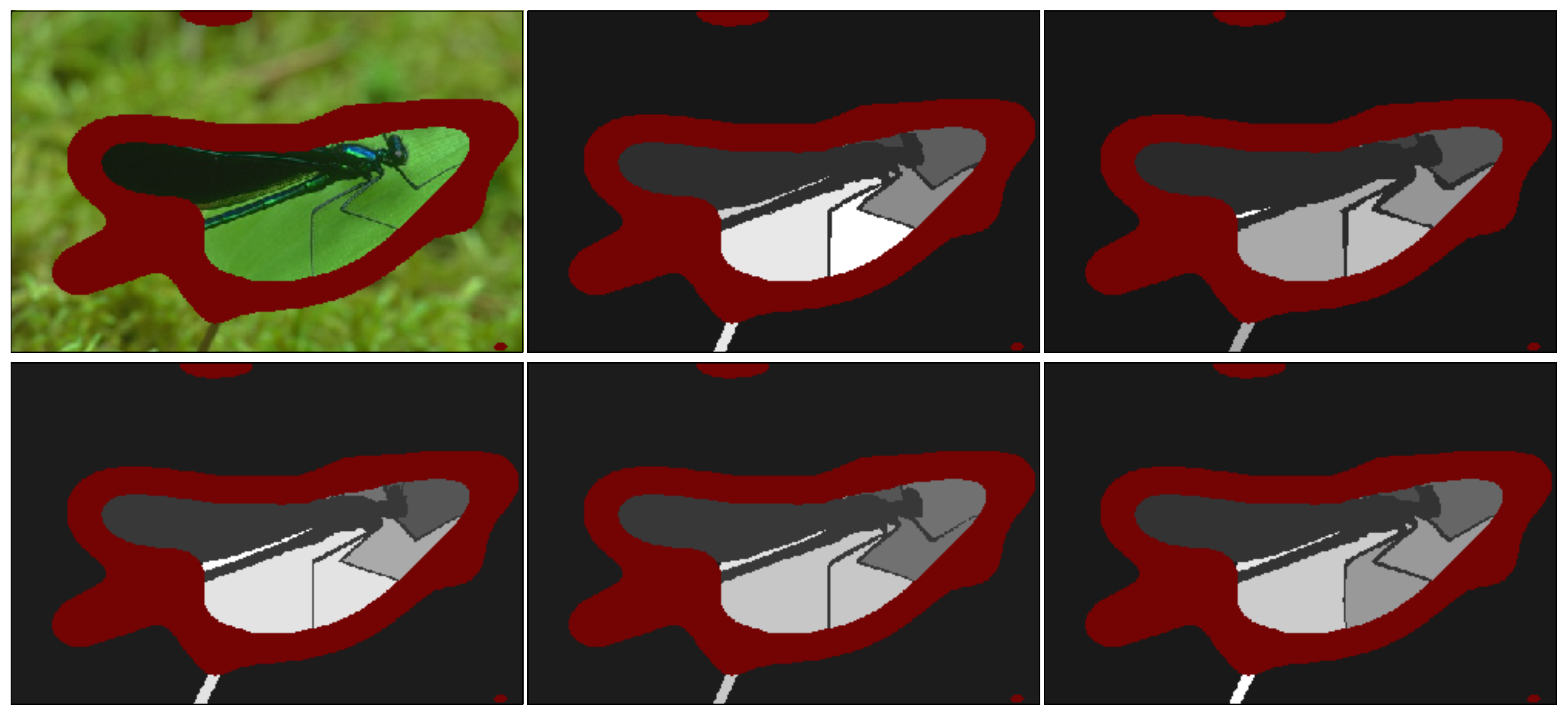}}{Examples of areas of high (left) and low (right) variability in human segmentation of natural images. In both groups, the top left image is the original, the next 5 images are segmentations by different humans. Areas masked in red have opposite uncertainty (left: low; right: high).}{fig:entropy-ex}

\section{Conclusion}

We have proposed a probabilistic model of segmentation with the goal of guiding future quantitative studies of human segmentation of natural images. The model follows from the probabilistic brain hypothesis and the normative approach that inference of visual features is based on probabilities that reflect natural environment statistics. Specifically, we showed that the Student-t mixture model that accounts for natural image statistics, performs better than the Gaussian mixture model, and goes a long way in closing the gap with state of the art, non--probabilistic algorithms. The probabilistic framework allowed us to include a regularization in the form of a prior that favors grouping of nearby pixels, while also exactly respecting reliability--based weighting of the likelihood and prior. Such a prior can be conceptualized as an effect from contextual knowledge provided by nearby visual features, and could therefore be linked to lateral interactions between cortical neurons. Lastly, our proposed model offered novel insight into the variability of human segmentation maps, suggesting that it may reflect uncertainty due to image ambiguity, which we found to be particularly prominent nearby the boundaries of different segments.

\bibliography{../references}
\end{document}